%% file: main.tex
\documentclass{article}
\usepackage[T1]{fontenc}
\usepackage[utf8]{inputenc}
\usepackage{freechroma_preprint,times}
\usepackage{xcolor}
\newcommand{\method}{FreeCHROMA}
\newcommand{\tbd}{\textit}
\input{math_commands.tex}

\usepackage{url}
\usepackage{booktabs}
\usepackage{graphicx}
\usepackage{float}
\usepackage{placeins}
\usepackage{hyperref}
\usepackage{cleveref}
\crefname{figure}{Fig.}{Figs.}
\Crefname{figure}{Fig.}{Figs.}
\crefname{table}{Tab.}{Tabs.}
\Crefname{table}{Tab.}{Tabs.}

\newcommand{\projecturl}{https://jiayi-wu-leo.github.io/freechroma/}
\newcommand{\papertitle}{FreeCHROMA: Generative Chroma Keying from Unaligned Background References}
\title{\papertitle}
\input{authors.tex}
\date{}
\hypersetup{
  pdftitle={\papertitle},
  pdfauthor={\paperauthorsplain},
  pdfsubject={Generative video compositing},
  pdfkeywords={video generation, chroma keying, foreground-background harmonization},
  colorlinks=true,
  linkcolor=blue,
  citecolor=blue,
  urlcolor=blue,
  bookmarksnumbered=true
}

\begin{document}
\maketitle

\begin{abstract}
In film production, chroma keying composites an approved foreground performance into a new environment, but conventional workflows require coordinated background plates and dedicated harmonization.
Recent generative methods often require pre-aligned backgrounds, focus on background-conditioned foreground relighting, or represent the environment only through an edited first frame.
These restrictions limit background selection, scene composition, and camera motion while adding preparation or editing effort.
We present FreeCHROMA, which adapts a pretrained video diffusion model in a parameter-efficient manner for generative video compositing from one to three unaligned background references.
The model integrates the foreground into the referenced environment, preserving foreground identity, motion, and fine visual details while generating a spatiotemporally coherent background.
Foreground–background harmonization occurs within the same generation process, without a separate relighting stage.
Optional text instructions help disambiguate complex spatial relationships and camera behavior.
To train the model, we construct pseudo-paired data from readily available real monocular videos through foreground–background separation and augmentation with existing foundation models.
The original videos serve as training targets, while automated processing and quality verification enable scalable data construction.
Across reference fidelity, foreground preservation, scene integration, harmonization, and overall video quality, FreeCHROMA consistently outperforms recent compositing and conditional video-generation baselines, with particularly clear gains under camera motion.
FreeCHROMA relaxes the constraints of existing tools to enable end-to-end generative chroma keying from unaligned inputs, with exploratory experiments demonstrating its potential in robotics.

\end{abstract}

\input{src/Intro}
\input{src/Related_Works}
\input{src/Method}
\input{src/Results}
\input{src/Conclusion}
\clearpage

\bibliography{references}
\bibliographystyle{freechroma}
\appendix
\input{src/Appendix}
\end{document}

%% file: math_commands.tex
\usepackage{amsmath,amsfonts,bm}

\def\eqref#1{equation~\ref{#1}}

\def\1{\bm{1}}

\DeclareMathAlphabet{\mathsfit}{\encodingdefault}{\sfdefault}{m}{sl}
\SetMathAlphabet{\mathsfit}{bold}{\encodingdefault}{\sfdefault}{bx}{n}



%% file: authors.tex
\author{%
  Jiayi Wu\textsuperscript{1}\quad
  Tianfu Wang\textsuperscript{1}\quad
  Rui Zhao\textsuperscript{2}\quad
  Florian Schiffers\textsuperscript{2}\quad
  Zhimin Chen\textsuperscript{2}\\[0.4ex]
  Cornelia Fermuller\textsuperscript{1}\quad
  Yezhou Yang\textsuperscript{3}\quad
  Christopher Metzler\textsuperscript{1}\quad
  Yiannis Aloimonos\textsuperscript{1}\\[0.7ex]
  {\small
    \textsuperscript{1}University of Maryland, College Park\quad
    \textsuperscript{2}Amazon\thanks{This work is unrelated to the authors’ positions at Amazon.}\quad
    \textsuperscript{3}Arizona State University}\\[0.7ex]
  {\small Project page: \href{\projecturl}{\nolinkurl{jiayi-wu-leo.github.io/freechroma/}}}%
}
\newcommand{\paperauthorsplain}{Jiayi Wu, Tianfu Wang, Rui Zhao, Florian Schiffers, Zhimin Chen, Cornelia Fermuller, YZ Yang, Christopher Metzler, Yiannis Aloimonos}

%% file: src/Intro.tex
\section{Introduction}
\label{sec:introduction}

Film and virtual-production workflows often capture an actor or object separately and composite the footage into an independently created environment. At this stage, the foreground performance---including identity, expression, motion, costume, and fine appearance---has usually been directed and approved. The goal is therefore not unconstrained video editing. It is to preserve that performance while constructing a plausible environment around it and changing only those appearance factors needed for coherent integration.

\begin{figure*}[t]
  \centering
  \includegraphics[width=\textwidth]{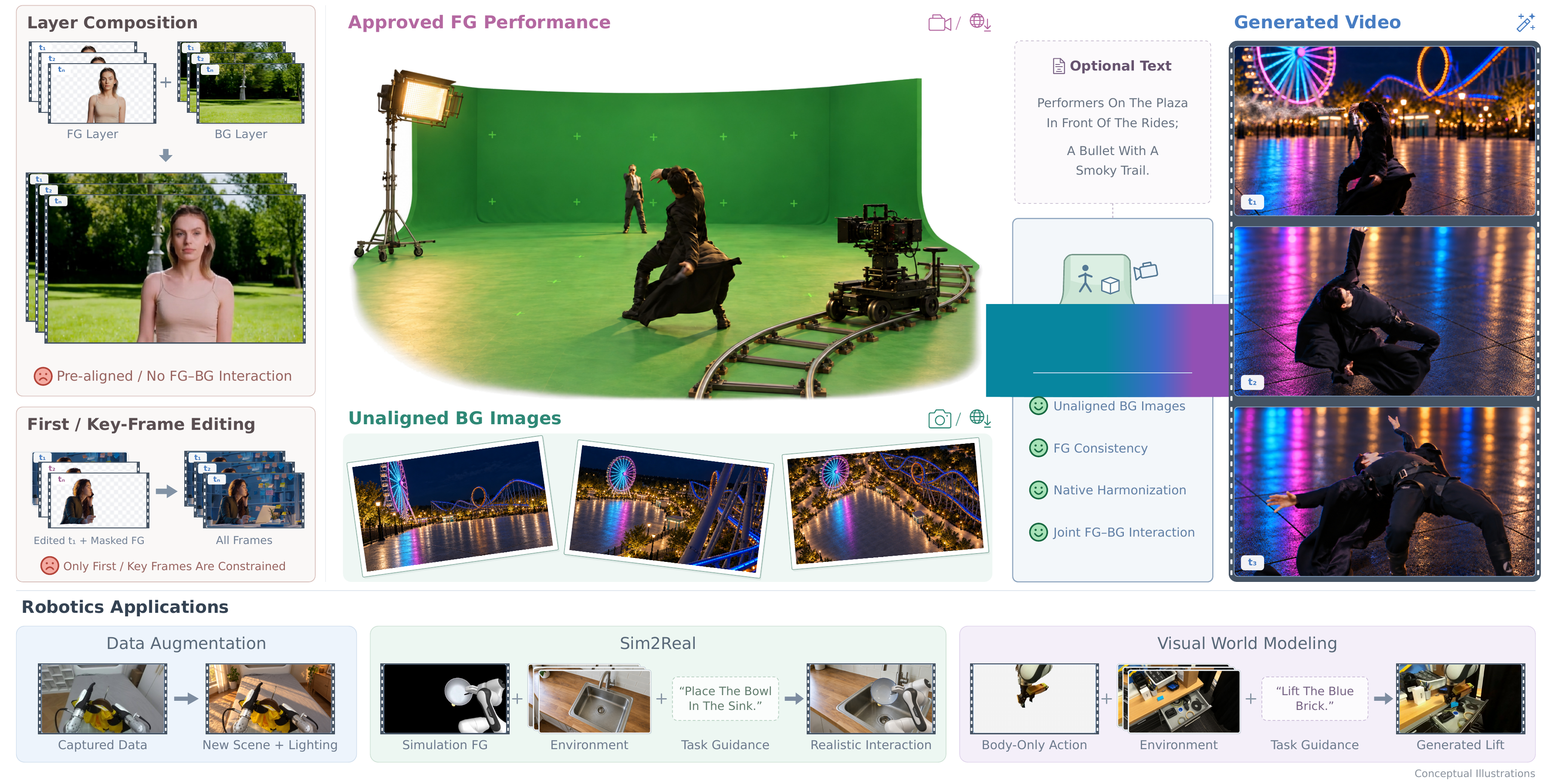}
  \caption{\textbf{FreeCHROMA and the production constraints it relaxes.} \textbf{Top left:} Layer composition requires pre-aligned videos and primarily relights the foreground against a fixed background. First/key-frame editing adds an image-editing step and constrains the environment only in selected views. \textbf{Top center and right:} FreeCHROMA preserves the foreground performance while using one or more unaligned background reference images throughout generation, jointly inferring composition, interactions, camera-aware background evolution, and harmonization, with optional text guidance. \textbf{Bottom:} Masked inputs retain all foreground elements for environment augmentation, the robot and interacting objects for sim-to-real transfer, or only the robot body for visual world modeling.
  }
  \label{fig:teaser}
\end{figure*}

Conventional chroma keying separates foreground and background effectively, but transfers much of the difficulty to production: background plates must agree with the foreground in viewpoint, framing, camera motion, perspective, and illumination, after which matting refinement and harmonization are still required~\citep{smith1996blue,sengupta2020background}. 
Recent generative systems reduce manual work but retain restrictive assumptions (Fig.~\ref{fig:teaser}, top left). Layer composition requires pre-aligned foreground and background videos and primarily relights an independent foreground layer against a fixed background, making it less suited to substantial camera motion or scene interaction. First/key-frame editing relies on an image editor to composite and harmonize selected frames before video generation. The environment is constrained only in those edited views, leaving newly revealed regions prone to drift; independently edited keyframes can also be inconsistent.

These restrictions are especially visible in mid-shot action sequences, such as fight choreography, parkour, or superhero-style stunts. Such shots commonly combine substantial subject motion with tracking, panning, or orbiting cameras, while retaining enough of the environment for foreground--background spatial relationships to matter. They therefore require more than relighting an approximately independent foreground layer: the environment must remain controllable as new regions enter the field of view, and the approved choreography must remain unchanged. Unified video editing models offer greater flexibility in instruction and reference conditioning, but often struggle to maintain fidelity to the supplied references.

We introduce \textbf{\method{}}, a \emph{generative chroma keying} method that uses unaligned background references and harmonizes foreground and background during generation. It takes a masked foreground video and its mask, one to three images of the desired environment, and optional text. The reference images need not match any output frame in camera pose, crop, or layout. The model learns a joint spatiotemporal composition: where the foreground belongs in the referenced environment, how the background evolves with foreground motion and implied camera behavior, and how foreground appearance adapts for natural integration. Optional text primarily helps resolve ambiguity, guiding generation toward the intended interpretation when the visual inputs admit multiple plausible compositions.

The central challenge is the scarcity of paired supervision. We construct pseudo-pairs from ordinary videos by automatically extracting foregrounds and perturbing their illumination, while creating clean, unaligned background references through foreground removal and viewpoint transformation of selected keyframes. Reconstructing the original video from these inputs jointly teaches reference-conditioned background generation and native harmonization, without green-screen capture, 3D annotations, or a separate relighting network.

We implement this formulation through lightweight adaptation of a Wan-family video diffusion transformer~\citep{wan2025}, incorporating foreground conditioning, persistent reference context, and optional text. We train the added input-projection weights for foreground and mask conditioning, the background-reference convolution, and LoRA adapters~\citep{hu2022lora}. Our main contribution is a scalable learning formulation that turns diverse natural videos into supervision for flexible composition.

\begin{samepage}
Our specific contributions are:
\begin{itemize}
    \item We formulate performance-preserving generative chroma keying with one to three spatially unaligned background references, learning composition rather than requiring it to be specified by a target plate.
    \item We introduce a pipeline for constructing pseudo-paired training data from readily available real monocular videos. It combines foreground--background separation with augmentation supported by existing foundation models, while retaining the original videos as training targets. Automated processing and quality verification enable scalable data construction.
    \item We integrate background synthesis and video-level harmonization within one parameter-efficient video diffusion model, without an explicit or post-hoc relighting stage. Across diverse foreground--background interactions and camera motions, our method outperforms state-of-the-art baselines in foreground preservation, background reference fidelity, and harmonization.
\end{itemize}
\par
\end{samepage}

%% file: src/Related_Works.tex
\section{Background and Related Work}
\label{sec:related}

Matting and keying recover foreground layers~\citep{smith1996blue,lin2021real}, while image harmonization and relighting adjust foreground appearance within an established composition~\citep{cong2020dovenet,zhang2025scaling}. Our setting additionally requires inferring scene composition and camera-dependent background evolution from unaligned background references.

\paragraph{Video relighting and background replacement.}
Recent open-source diffusion systems extend relighting and harmonization to video. RelightVid incorporates temporal modeling into background-conditioned foreground relighting~\citep{fang2025relightvid}, while Light-A-Video constructs a training-free video relighting pipeline from pretrained image relighting and video-generation models~\citep{zhou2025light}. Lumen generates relit foreground videos conditioned on textual descriptions of the desired lighting and background~\citep{zeng2025lumen}. IllumiCraft further introduces geometry-aware conditioning for controllable and temporally consistent video illumination~\citep{lin2025illumicraft}. AnyPortal and FlowPortal combine video background replacement with temporally coherent foreground relighting in zero-shot or training-free formulations~\citep{gao2025anyportal,gao2026flowportal}. These methods demonstrate that pretrained generative models can coordinate foreground appearance and illumination over time. However, this formulation can be viewed as a simplification of the full compositing problem: it largely ignores explicit foreground–background interaction and instead treats the input background sequence—often already manually aligned or even left unaligned—as a proxy light source or global conditioning signal, rather than as a fully structured, spatially consistent environment to be jointly inferred. PAI-Studio~\citep{gao2026paistudiocinematicvideobackground} associates background references with designated keyframes, implicitly assuming that the references are pre-aligned with the foreground at those frames. \method{} instead treats one or more spatially unaligned background images as persistent video-level references and jointly synthesizes the environment, its spatiotemporal relationship to the foreground performance, and the appearance changes required for integration.

\paragraph{Conditional video generation and keyframe-guided restylization.}
Latent diffusion models~\citep{rombach2022high} and diffusion transformers~\citep{peebles2023scalable} provide scalable backbones for conditional video synthesis. 
VACE~\citep{jiang2025vace} and Bernini~\citep{bernini} unify multiple video creation and editing tasks through spatially organized visual and textual conditions. Despite their versatility, these models often struggle to achieve precise foreground preservation and faithful adherence to background references.
Closely related in production motivation, ID-V2V performs identity-preserving video restylization from a source video and a stylized first frame, optionally supplemented with additional keyframes, propagating the specified scene, lighting, and appearance across the sequence while preserving the captured performance~\citep{xu2026id}. This keyframe-based cascade offloads foreground--background composition to an upstream image editor, making the final video dependent on the spatial and appearance consistency of the edited keyframes and effectively using the downstream video model primarily as a controllable temporal interpolator rather than as the compositor itself. More importantly, the target environment is represented only within the field-of-view covered by the edited keyframes; regions newly revealed by foreground or camera motion remain underconstrained and may drift away from the intended scene. \method{} addresses a distinct control regime: it composes a preserved foreground performance directly with one to three standalone background references that need not align with the source video or any output frame. These references remain available throughout generation, enabling the model to jointly infer composition, background evolution, and foreground--background harmonization without a separately edited first frame or an additional relighting stage.

%% file: src/Method.tex
\section{Method}
\label{sec:method}

\subsection{Problem formulation}
Let $F\in\mathbb{R}^{T\times H\times W\times3}$ be a foreground-only video with alpha mask $M$, let $\mathcal{R}=\{R_k\}_{k=1}^{K}$ contain $K\in\{1,2,3\}$ images of a desired environment, and let $c$ be optional text. We seek
\begin{equation}
  \widehat V = G_{\theta}(F,M,\mathcal{R},c),
  \label{eq:task}
\end{equation}
where $\widehat V$ preserves the semantic and temporal content of $F$ while generating and harmonizing a background consistent with $\mathcal{R}$. The references are not target frames: no pixel correspondence, camera calibration, or pose alignment is assumed. The output is consequently not unique. Our objective is conditional plausibility and reference fidelity under foreground preservation, rather than reconstruction of a particular unseen background trajectory.

\subsection{Scalable pseudo-pairs from natural video}
\label{sec:data-pipeline}

\begin{figure*}[!ht]
  \centering
  \includegraphics[width=0.99\textwidth]{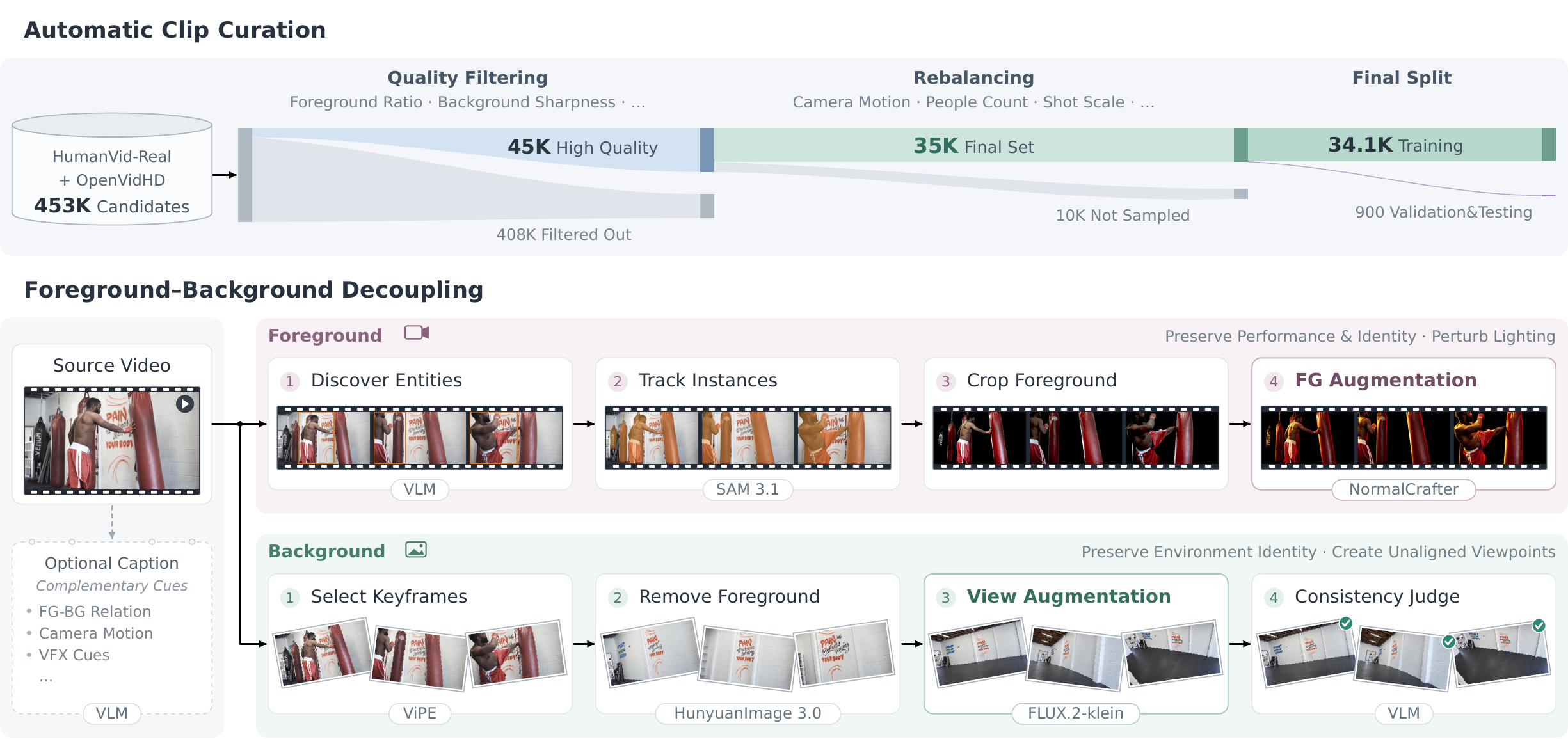}
  \caption{\textbf{Automatic pseudo-pair construction from natural videos.}
  \textbf{Clip curation:} A shot detector, conventional image-quality tests, and a vision--language model (VLM) retain clips with a visible foreground and informative background.
  \textbf{Foreground augmentation:} The VLM enumerates the foreground entities; SAM~3.1 tracks their masks $M$; and a temporally coherent photometric and normal-guided illumination transform produces $(\widetilde F,M)$.
  \textbf{Background reference construction:} A motion proxy selects one to three keyframes, a frozen image editor removes all selected entities, a frozen novel-view model changes viewpoint and crop, and a VLM quality gate retains scene-consistent references $\widetilde{\mathcal R}$.
  The conditions map to the untouched source video $V$, yielding scalable supervision without green-screen capture, manual compositing, or 3D annotation. Tool roles and selected models are labeled in the figure; further implementation details are provided in Appendix~\ref{app:data}. Visuals are schematic.}
  \label{fig:data-pipeline}
\end{figure*}

We derive supervision from an ordinary video $V$ by constructing
\begin{equation}
  \big(\widetilde F,M,\widetilde{\mathcal R},c\big) \longrightarrow V.
  \label{eq:pseudopair}
\end{equation}
The original video supplies a temporally coherent target in which foreground, background, camera motion, and illumination already agree. \Cref{fig:data-pipeline} summarizes the process; data-construction details are provided in Appendix~\ref{app:data}.

\paragraph{Video selection and foreground extraction.}
We retain short clips in which both the salient foreground and surrounding environment are sufficiently visible and sharp. A vision-language model enumerates foreground entities, including interacting objects that should remain attached to the performance. SAM~3.1, a released refinement of SAM~3~\citep{carion2026sam}, segments and tracks those concepts over time. We take the union of the selected instance masks,
and form $F=M\odot V$. Clips with unstable masks, excessive foreground occupancy, shot changes, or negligible background evidence are rejected.

\paragraph{Foreground illumination augmentation.}
The inexpensive construction $F=M\odot V$ introduces a degenerate shortcut: on the matte support, the conditioning foreground is identical to the regression target, so copying $F$ is optimal regardless of the referenced environment. This shortcut fails at inference, where the foreground and target environment are generally captured under different illumination. We mitigate it by perturbing only the conditioning path while leaving the target video unchanged:
\begin{equation}
  \widetilde F=T_{\theta}(F;N), \qquad \theta\sim p(\theta),
  \label{eq:illumination}
\end{equation}
where $N$ is an optional frame-aligned normal map. $T_{\theta}$ combines global photometric changes, low-frequency spatial variation, and normal-guided directional shading; all parameters are shared across the clip and the original matte support is restored after transformation. Because $(\widetilde F,\widetilde{\mathcal R},c)$ must still predict the unmodified $V$, an exact foreground copy is no longer optimal. The model must instead preserve performance content while using the referenced environment to infer a compatible appearance. This converts natural-video supervision into an explicit harmonization signal without requiring paired relit foregrounds. The transform is deliberately bounded and designed to resemble illumination variation rather than arbitrary corruption; full operators and sampling ranges are given in Appendix~\ref{app:illum-aug}.

\paragraph{Unaligned background references.}
We use camera trajectories estimated by ViPE~\citep{huang2025vipe} to assess motion magnitude and select one to three source keyframes $B_k$ accordingly. The VLM foreground labels and masks guide a frozen image-editing model, $\operatorname{RemoveFG}$, which removes the selected entities from $B_k$. A frozen viewpoint-generation model then changes the cleaned background's viewpoint and crop:
\begin{equation}
  \widetilde R_k=A_{\mathrm{view}}\!\left(\operatorname{RemoveFG}(B_k)\right).
  \label{eq:viewaug}
\end{equation}
This transformation is essential because an unmodified keyframe can be copied or aligned directly to its source frame. View augmentation forces the references to encode environment identity, appearance, and partial geometry rather than a predetermined composition. We reject transformations that change the underlying scene or introduce conspicuous artifacts using VLM-based scene-consistency and quality checks. The reference-generation models and quality checks are described in Appendix~\ref{app:data_bg} (Background references).

\subsection{Joint composition and harmonization}
\label{sec:model}

\begin{figure*}[!t]
  \centering
  \includegraphics[width=\textwidth]{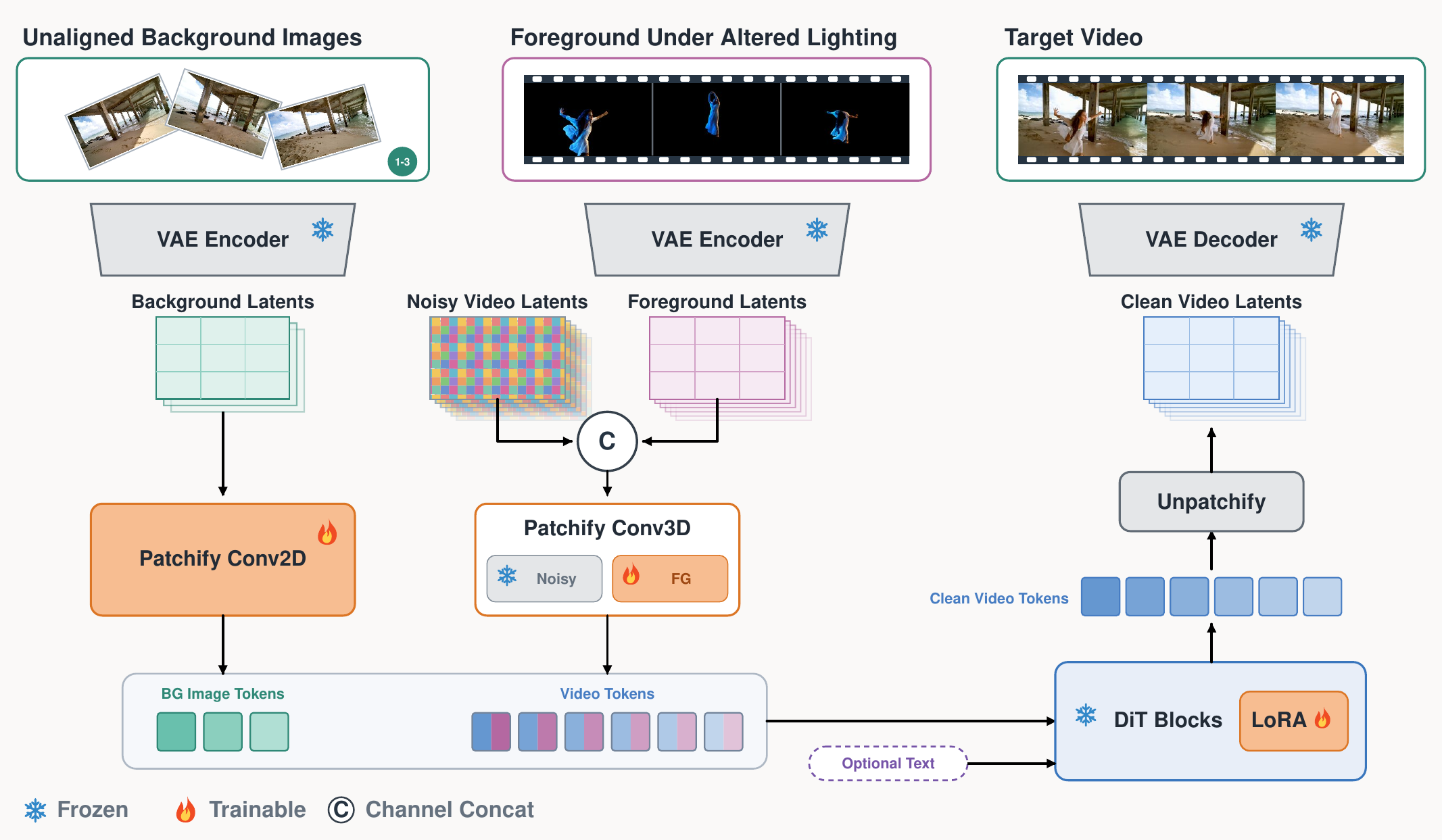}
  \caption{\textbf{Parameter-efficient FreeCHROMA architecture.} During training, a frozen VAE encodes the target and foreground streams. The noisy target latent, foreground latent, and aligned mask condition are concatenated channel-wise and enter through the expanded patchify projector in \cref{eq:patchify}; background reference tokens are produced by a trainable convolution and prepended to the video sequence; optional text uses the pretrained cross-attention path. Flow-matching supervision updates the added foreground and mask input weights, the background-reference convolution, and DiT LoRA adapters. At inference, the same conditioned denoiser transforms noise into a complete video, which the frozen VAE decodes. Scene synthesis, spatiotemporal composition, and foreground harmonization occur in one denoising process, without an auxiliary relighting or frame-wise harmonization network.}
  \label{fig:architecture}
\end{figure*}

We initialize from Wan2.2-Fun-A14B-Control~\citep{wan2025} and fine-tune its high- and low-noise experts separately; implementation details are provided in Appendix~\ref{app:implementation}. \Cref{fig:architecture} summarizes the conditioning architecture and marks the frozen and trainable components. A frozen VAE encodes the target $V$ and foreground condition $\widetilde F$ as $z_0$ and $z_F$, respectively. At diffusion time $\tau$, noise gives $z_\tau$. We concatenate the noisy-target latent, foreground latent, and explicit mask condition channel-wise and map them jointly into transformer tokens through an expanded patchify input projection:
\begin{equation}
 h_0=P_{\mathrm{patch}}\!\left([z_\tau\,\Vert\,z_F\,\Vert\,m]\right),
 \label{eq:patchify}
\end{equation}
where $m$ denotes $M$ on the latent grid and $\Vert$ denotes channel concatenation. The pretrained target-latent weights remain frozen; only the added foreground and mask weights in $P_{\mathrm{patch}}$ are trained. This dense interface preserves foreground--target alignment at every space--time position without a separate foreground encoder.

Each reference is encoded by the frozen VAE and mapped to tokens $E_k$ by a trainable background-reference convolution. We concatenate $E_{1:K}$ and prepend them to the noisy video-token sequence as persistent visual context. Unlike a first-frame image condition, these tokens remain accessible to all denoising blocks and output frames. Text tokens $E_c$ enter through the pretrained cross-attention pathway when an instruction is present; classifier-free dropout over text and references supports absent text and robust variable-cardinality conditioning.

We follow the backbone's flow-matching parameterization. For sampled $\tau$ and noise $\epsilon$, the trainable parameters minimize
\begin{equation}
 \mathcal L_{\mathrm{FM}}=
 \mathbb E_{V,\tau,\epsilon}
 \left[\left\|v_{\theta}(z_\tau,h_0,E_{1:K},E_c,\tau)-v^{\star}\right\|_2^2\right],
 \label{eq:objective}
\end{equation}
with the standard target velocity $v^{\star}$ defined by the selected Wan scheduler. The VAE, text encoder, and original DiT weights remain frozen. We optimize the added foreground and mask input-projection weights, the background-reference convolution, and DiT LoRA adapters~\citep{hu2022lora}. No auxiliary foreground relighting network, per-frame harmonizer, explicit depth loss, or reconstruction model is used at inference.

%% file: src/Results.tex
\section{Experimental Results}

\subsection{Evaluation Protocol}
We evaluate FreeCHROMA on 500 foreground videos (167 HumanVid-Real~\citep{DBLP:conf/nips/00010ZF0LTCX0L24} and 333 OpenVidHD~\citep{nan2024openvid}) spanning static to large camera motion. Each foreground video is paired with two different sets of unaligned background references, yielding 1,000 test cases. All methods are evaluated on the same foreground--background pairs, with inputs adapted to their supported conditioning modalities.

Baselines comprise Lumen~\citep{zeng2025lumen}, FlowPortal~\citep{gao2026flowportal}, IllumiCraft~\citep{lin2025illumicraft}, two ID-V2V variants (Paste and FLUX.2)~\citep{xu2026id}, VACE~\citep{jiang2025vace}, and Bernini~\citep{bernini}. Visually conditioned methods receive the background image; Lumen receives only a deterministic Qwen caption of it. We assess video quality with VBench~\citep{huang2023vbench} and VBench-2.0~\citep{zheng2025vbench2}, compositing with a vision-language evaluator, and camera-trajectory consistency with BA-Track~\citep{chen2025back}.

\subsection{General Video Quality}
Tab.~\ref{tab:vbench} shows that FreeCHROMA achieves strong overall video quality, outperforming most baselines across the evaluated metrics. These results suggest that our pipeline effectively preserves the pretrained backbone's general video generation quality while adapting it to generative compositing.
\begin{table*}[!htbp]
\centering
\caption{
Video quality on the evaluation benchmark: VBench~\citep{huang2023vbench}, except multi-view consistency from VBench-2.0~\citep{zheng2025vbench2}. Best results are bold; second-best are underlined. \textsuperscript{\textdagger}Lumen receives a caption; other methods use visual background references.
}
\label{tab:vbench}
\small
\setlength{\tabcolsep}{5.5pt}
\renewcommand{\arraystretch}{1.08}
\resizebox{0.95\linewidth}{!}{%
    \begin{tabular}{@{}lccccccc@{}}
        \toprule
        & \multicolumn{2}{c}{Content Consistency}
        & \multicolumn{3}{c}{Temporal Quality}
        & \multicolumn{2}{c}{Visual Quality} \\
        \cmidrule(lr){2-3}
        \cmidrule(lr){4-6}
        \cmidrule(lr){7-8}

        Method
        & \shortstack{Subject\\Cons. $\uparrow$}
        & \shortstack{Background\\Cons. $\uparrow$}
        & \shortstack{Motion\\Smoothness $\uparrow$}
        & \shortstack{Temporal\\Flickering $\uparrow$}
        & \shortstack{Multi-view\\Cons. $\uparrow$}
        & \shortstack{Aesthetic\\Quality $\uparrow$}
        & \shortstack{Imaging\\Quality $\uparrow$} \\
        \midrule

        Lumen~\citep{zeng2025lumen}\textsuperscript{\textdagger}
        & \underline{94.26} & 93.22 & 98.06 & 96.14 & 19.06 & 53.82 & 71.25 \\

        VACE~\citep{jiang2025vace}
        & 93.94 & 93.28 & 97.73 & 95.98 & 24.56 & {57.33} & 72.42 \\

        IllumiCraft~\citep{lin2025illumicraft}
        & 91.64 & 92.12 & 96.87 & 95.78 & 15.07 & 53.23 & 70.61 \\

        FlowPortal~\citep{gao2026flowportal}
        & 93.23 & 92.53 & 97.77 & 95.95 & 15.63 & 53.85 & 67.24 \\

        ID-V2V + Paste~\citep{xu2026id}
        & 91.70 & 93.18 & 98.07 & 96.11 & 25.01 & 56.49 & 73.30 \\

        ID-V2V + FLUX.2 ~\citep{xu2026id}
        & 93.14 & 92.64 & 97.92 & 95.87 & \underline{27.36} & 57.64 & \textbf{73.68} \\

        Bernini ~\cite{bernini}
        & 94.08 & \textbf{94.57} & \underline{98.31} & \underline{96.33} & \textbf{28.78} & \textbf{59.40} & {73.30} \\
        \midrule

        \textbf{FreeCHROMA}
        & \textbf{94.59}
        & \underline{93.65}
        & \textbf{98.55}
        & \textbf{97.16}
        & {26.51}
        & \underline{57.82}
        & \underline{73.63} \\

        \bottomrule
    \end{tabular}%
}
\end{table*}

\subsection{Foreground Preservation and Scene Integration}
Following the evaluation criteria of PAI-Studio~\citep{gao2026paistudiocinematicvideobackground}, we assess foreground preservation, edge quality, foreground--background fusion, and illumination harmony using a vision-language model (Tab.~\ref{tab:task_metrics}); the evaluation protocol is detailed in Appendix~\ref{app:vlm}. FreeCHROMA achieves the highest score on all four criteria. These results suggest that joint generation preserves foreground fidelity while adapting appearance and spatial relationships to the target environment.

\begin{table*}[!htbp]
\centering
\caption{
Task-specific evaluation on the novel-background benchmark.
VLM scores measure foreground preservation,
boundary quality, scene integration, and illumination harmony.
Scores are averaged over 1,000 generated videos (500 foreground videos, each paired with two different sets of background references). Best results are bold; second-best are underlined.
}
\label{tab:task_metrics}
\small
\setlength{\tabcolsep}{5pt}
\renewcommand{\arraystretch}{1.08}
\resizebox{0.8\linewidth}{!}{%
\begin{tabular}{@{}lcccc@{}}
\toprule
& \multicolumn{2}{c}{Foreground Quality}
& \multicolumn{2}{c}{Scene Integration} \\
\cmidrule(lr){2-3}
\cmidrule(lr){4-5}

Method
& \shortstack{FG Preservation\\VLM $\uparrow$}
& \shortstack{Edge Quality\\VLM $\uparrow$}
& \shortstack{FG--BG Fusion\\VLM $\uparrow$}
& \shortstack{Illum. Harmony\\VLM $\uparrow$} \\
\midrule

Lumen~\citep{zeng2025lumen}\textsuperscript{\textdagger}
& 9.778 & 6.722 & 6.470 & 7.040 \\

VACE~\citep{jiang2025vace}
& 9.852 & \underline{7.840} & \underline{7.580} & \underline{7.864} \\

IllumiCraft~\citep{lin2025illumicraft}
& 8.330 & 6.196 & 6.278 & 6.932 \\

FlowPortal~\citep{gao2026flowportal}
& 5.624 & 5.842 & 5.802 & 6.450 \\

ID-V2V + Paste ~\citep{xu2026id}
& 9.636 & 6.304 & 6.372 & 6.088  \\

ID-V2V + FLUX.2~\citep{xu2026id}
& \underline{9.901} & 7.386 & 7.058 & 7.602 \\

Bernini~\citep{bernini}
& 9.180 & 6.586 & 6.476 & 7.130 \\
\midrule

\textbf{FreeCHROMA}
& \textbf{9.904} & \textbf{8.172} & \textbf{7.728} & \textbf{7.956} \\

\bottomrule
\end{tabular}%
}
\end{table*}

\begin{figure*}[!t]
  \centering
  \includegraphics[width=\textwidth]{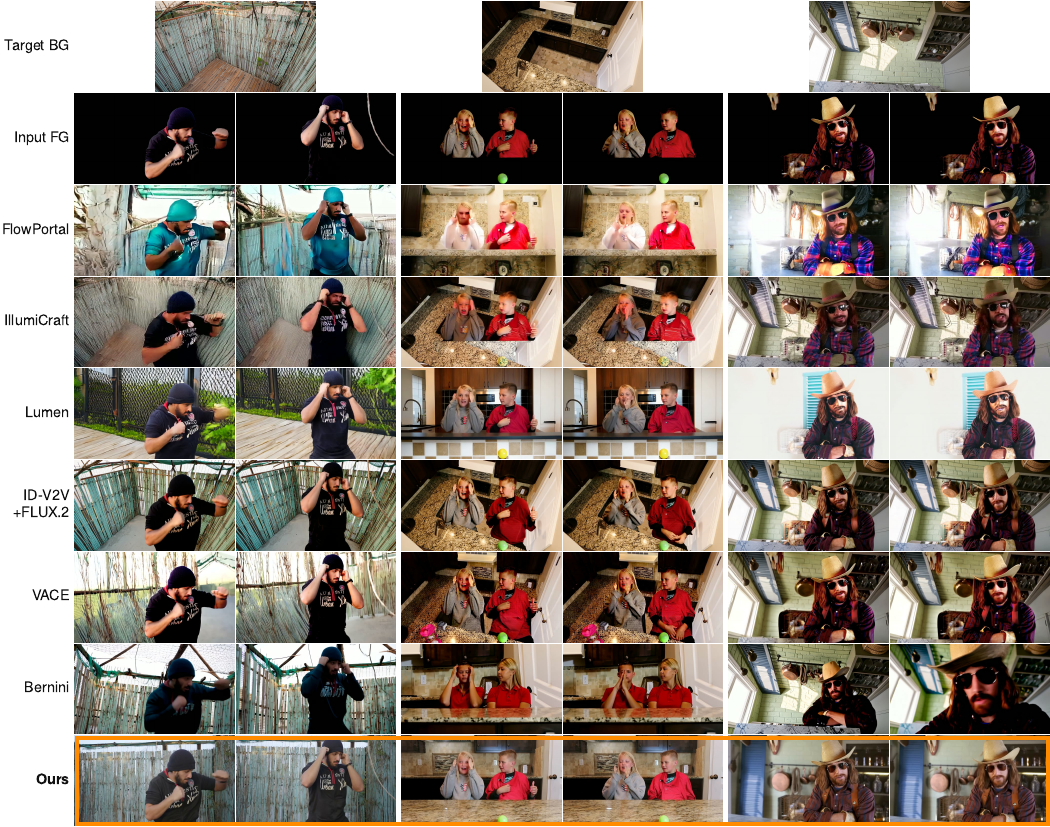}
  \caption{\textbf{Qualitative comparison of generated videos.} Given background reference images and a masked foreground sequence, each method generates an output video; two frames sampled at matched relative times are shown. FreeCHROMA faithfully preserves the foreground performance while producing more plausible foreground--background spatial relationships, including viewpoint consistency, relative scale, and physical support. It also generates more natural camera motion and harmonizes illumination and appearance across foreground and background. \textcolor{blue}{See the \href{\projecturl}{project page} for more video examples.}
    }
  \label{fig:qual_compare}
\end{figure*}

Fig.~\ref{fig:qual_compare} further shows that FreeCHROMA achieves foreground--background coherence in spatial geometry and lighting, preserves foreground appearance, motion, and fine-grained details such as text and logos on clothing, and maintains high background fidelity and consistency across viewpoints.

\subsection{Camera-Trajectory Consistency}
A convincing composite requires the synthesized environment to remain geometrically coherent with the foreground and follow its camera motion. We therefore compare each generated video with the original source video from which its foreground input was extracted. Camera trajectories are estimated using BA-Track~\citep{chen2025back} and aligned with a per-video similarity transform.

\begin{table}[H]
\centering
\caption{
Camera-trajectory consistency between generated videos and the original source videos from which the foreground inputs are extracted. Best results are bold; second-best are underlined.}
\label{tab:camera_trajectory}
\small
\setlength{\tabcolsep}{5pt}
\renewcommand{\arraystretch}{1.08}
\resizebox{0.88\linewidth}{!}{%
\begin{tabular}{@{}lccccc@{}}
\toprule
& \multicolumn{2}{c}{Global Trajectory Error}
& \multicolumn{2}{c}{Local Motion Error}
& \multicolumn{1}{c}{Motion Extent} \\
\cmidrule(lr){2-3}
\cmidrule(lr){4-5}
\cmidrule(lr){6-6}

Method
& \shortstack{ATE\\$\downarrow$}
& \shortstack{ATE/Path (\%)\\$\downarrow$}
& \shortstack{RPE Translation\\$\downarrow$}
& \shortstack{RPE Rotation ($^\circ$)\\$\downarrow$}
& \shortstack{Relative Path Length\\$\rightarrow 1$} \\
\midrule

Lumen~\citep{zeng2025lumen}\textsuperscript{\textdagger}
& 0.015 & 9.13 & 0.008 & 0.593 & 0.834 \\

VACE~\citep{jiang2025vace}
& \underline{0.006} & \underline{4.50} & \underline{0.005} & \underline{0.176} & \underline{0.936} \\

IllumiCraft~\citep{lin2025illumicraft}
& 0.017 & 9.25 & 0.009 & 0.653 & 0.823 \\

FlowPortal~\citep{gao2026flowportal}
& 0.015 & 9.00 & 0.008 & 0.614 & 0.866 \\

ID-V2V + Paste~\citep{xu2026id}
& 0.011 & 6.11 & 0.006 & 0.192 & 0.929 \\

ID-V2V + FLUX.2~\citep{xu2026id}
& 0.009 & 5.74 & 0.006 & 0.201 & 0.902 \\

Bernini~\citep{bernini}
& 0.008 & 5.46 & \underline{0.005} & 0.253 & 0.893 \\

\midrule

\textbf{FreeCHROMA}
& \textbf{0.004}
& \textbf{3.96}
& \textbf{0.003}
& \textbf{0.119}
& \textbf{0.940} \\

\bottomrule
\end{tabular}%
}
\end{table}

Absolute Trajectory Error (ATE) measures global trajectory disagreement after per-video Sim(3) alignment, while ATE/Path normalizes ATE by the source
trajectory length. The translational and rotational components of Relative Pose Error (RPE) measure local frame-to-frame motion disagreement. Relative Path Length
is the ratio between the aligned generated and source trajectory
lengths, for which $1.0$ is ideal.
Tab.~\ref{tab:camera_trajectory} shows that FreeCHROMA achieves the lowest global and local trajectory errors and a relative path length closest to $1.0$, indicating more faithful preservation of the source camera motion.

\subsection{Ablation Study}
We remove each training-time augmentation separately and retrain for the same number of training steps with all other settings unchanged (Tab.~\ref{tab:ablation_vlm_vbench_camera}).
Removing foreground illumination perturbation lowers foreground--background fusion from $7.728$ to $6.384$ and illumination harmony from $7.956$ to $6.958$, suggesting that training with mismatched foreground lighting helps the model adapt foreground appearance to the target environment.
Qualitatively, this variant tends to preserve the source foreground lighting and color casts, resulting in a visible mismatch with the target environment (Fig.~\ref{fig:ablation_qual}).
Removing background viewpoint augmentation increases ATE from $0.004$ to $0.005$ and shifts relative path length from $0.940$ to $0.909$, indicating weaker agreement with the source camera motion.
Qualitative results show that this variant tends to inherit the unaligned reference viewpoint, producing inconsistent scene geometry.
The full model performs best across all metrics in this ablation, supporting the combined use of the two augmentations for harmonization and camera-trajectory consistency.
\textcolor{blue}{Qualitative comparisons are given in Appendix~\ref{sec:app-ablation} (Fig.~\ref{fig:ablation_qual}), with video examples on the \href{\projecturl}{project page}.}

\begin{table*}[!htbp]
  \centering
  \caption{
  Ablation results on training augmentation components.
  }
  \label{tab:ablation_vlm_vbench_camera}
  \small
  \setlength{\tabcolsep}{5pt}
  \renewcommand{\arraystretch}{1.08}
  \resizebox{\linewidth}{!}{%
  \begin{tabular}{@{}lcccccc@{}}
  \toprule
  & \multicolumn{2}{c}{VLM Evaluation}
  & \multicolumn{2}{c}{VBench Evaluation}
  & \multicolumn{2}{c}{Camera Trajectory} \\
  \cmidrule(lr){2-3}
  \cmidrule(lr){4-5}
  \cmidrule(lr){6-7}

  Method
  & \shortstack{FG--BG Fusion\\$\uparrow$}
  & \shortstack{Illum. Harmony\\$\uparrow$}
  & \shortstack{Subject Cons.\\$\uparrow$}
  & \shortstack{Motion Smoothness\\$\uparrow$}
  & \shortstack{ATE\\$\downarrow$}
  & \shortstack{Relative Path Length\\$\rightarrow 1$} \\
  \midrule

  w/o FG illumination perturbation
  & 6.384 & 6.958 & 94.16 & 98.50 & 0.006 & 0.918 \\

  w/o BG viewpoint augmentation
  & 7.638 & 7.902 & 94.56 & 98.45 & 0.005 & 0.909 \\

  \midrule

  \textbf{FreeCHROMA}
  & \textbf{7.728} & \textbf{7.956} & \textbf{94.59}
  & \textbf{98.55} & \textbf{0.004} & \textbf{0.940} \\

  \bottomrule
  \end{tabular}%
  }
\end{table*}

\subsection{Downstream Robotics Applications}
\label{sec:robot}
\method{} preserves selected scene elements and their motion while coherently generating the rest. Beyond background augmentation of captured manipulation videos, varying the mask supports sim-to-real transfer by retaining the robot body and interacting objects, and visual world modeling by retaining only the robot body. We explore both settings on selected episodes from RoboCasa~\citep{robocasa365} and DROID~\citep{khazatsky2024droid} without robot-specific training. The examples show coherent motion and plausible interactions in the reference environments, including successful and failed outcomes. These results suggest that a compositing model can supply realistic manipulation videos and potentially serve as a rough visual world model. \textcolor{blue}{Examples and discussion are provided in Appendix~\ref{app:robotics} (Figs.~\ref{fig:sim2real_grid} and~\ref{fig:robot}), with video visualizations on the \href{\projecturl}{project page}.}

%% file: src/Conclusion.tex
\section{Conclusion and Limitations}
\label{sec:conclusion}
We presented \textbf{\method{}}, which preserves an existing foreground performance while jointly learning composition, background evolution, and foreground--background harmonization from one to three unaligned reference images. A scalable pseudo-pair pipeline built from natural videos and parameter-efficient adaptation of a pretrained video diffusion transformer enable flexible composition without pre-aligned plates, edited keyframes, or a separate relighting stage. Two limitations remain. The references provide context for short clips but no persistent scene memory for long-form generation. Independently generated shots also lack precise cross-shot foreground--background consistency beyond coarse textual guidance. Persistent environment representations and explicit cross-shot spatial constraints are promising directions for future work.

%% file: src/Appendix.tex
\providecommand{\tbd}[1]{\textbf{[TBD: #1]}}

\raggedbottom

\section{Qualitative Ablation}
\label{sec:app-ablation}

\begin{figure}[!t]
  \centering
  \includegraphics[width=\textwidth]{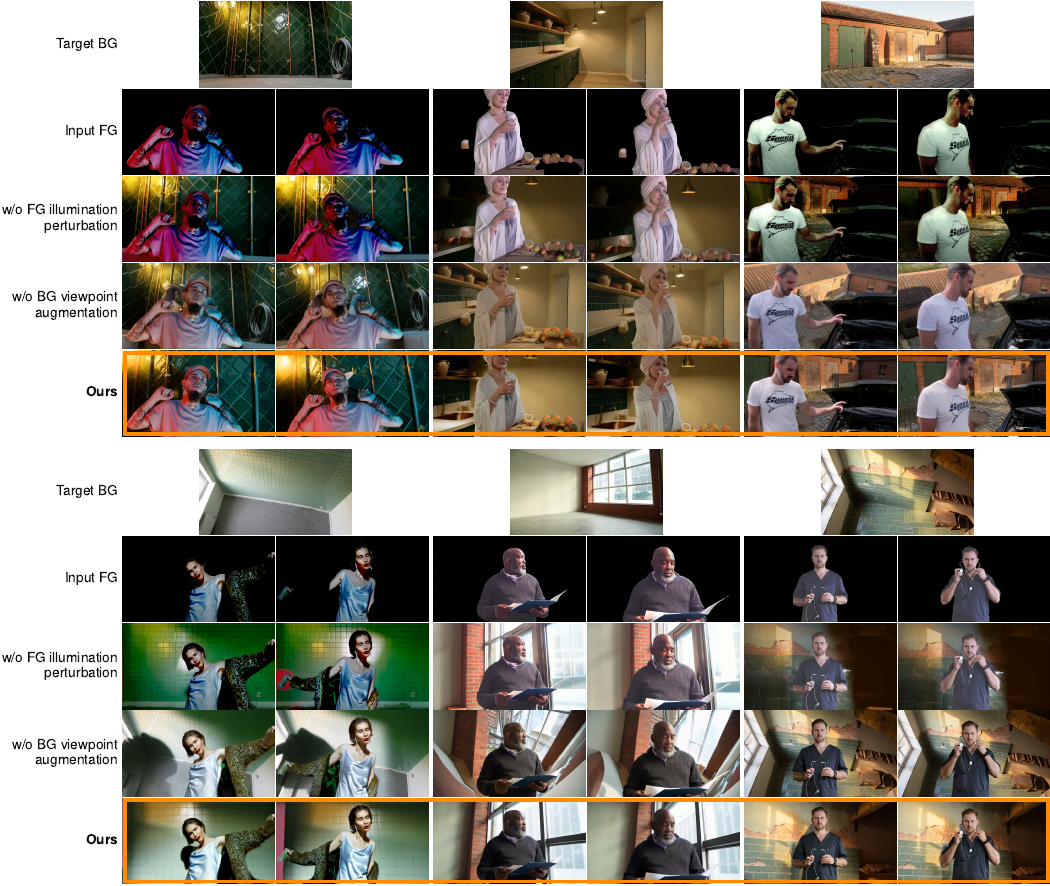}
  \caption{\textbf{Qualitative ablation of the two training-time augmentations.}
  For each sample we show the environment reference (\emph{Target BG}), the masked foreground control (\emph{Input FG}), and two frames of the videos generated by the two ablated models and by the full model (\textbf{Ours}, orange).
  Without the foreground illumination perturbation, the model tends to copy the illumination of the input foreground instead of re-lighting it for the target environment.
  Without the background viewpoint augmentation, the model is trained only on pixel-aligned background plates; given an unaligned reference at test time, it tends to force the reference's viewpoint into the output and composites a scene that is inconsistent with the foreground.
  The full model re-lights the foreground to match the target environment and uses the reference as an unaligned appearance cue, producing a consistent scene. \textcolor{blue}{See the \href{\projecturl}{project page} for more video examples.}}
  \label{fig:ablation_qual}
\end{figure}

We complement the quantitative ablation in \cref{tab:ablation_vlm_vbench_camera} with qualitative results for the two training-time augmentations that make our model robust to real inputs. Both ablated models are trained with the same data and schedule as the full model, for the same number of training steps; only the ablated component is removed.

\paragraph{w/o FG illumination perturbation.}
The foreground illumination perturbation (photometric changes and normal-guided shading) is disabled, so the foreground control seen during training always carries the illumination of the target video. The model therefore learns that the output illumination can be read off the input and, at test time, tends to copy the illumination of the input foreground instead of re-lighting it for the target environment (\cref{fig:ablation_qual}, top row): stage lighting, color casts, and exposure of the source footage are carried over unchanged, and the composite does not look like it was captured in the target scene.

\begin{samepage}
\paragraph{w/o BG viewpoint augmentation.}
The background references are taken from the original camera view (pixel-aligned with the target video) instead of the viewpoint-augmented, unaligned plates described in Appendix~\ref{app:data_bg} (Background references). The model then learns to treat the reference as an aligned background layer. Given an unaligned reference at test time, it tends to force the reference's viewpoint into the output and composites a scene that is inconsistent with the foreground (\cref{fig:ablation_qual}, bottom row); the full model instead uses the reference as an unaligned appearance cue and synthesizes a background whose layout is consistent with the foreground camera.
\par
\end{samepage}

\paragraph{Discussion.}
The two augmentations address complementary failure modes. Without the illumination perturbation the background is plausible but the foreground is not harmonized; without the viewpoint augmentation the foreground is harmonized but the scene geometry breaks.

\FloatBarrier
\section{Downstream Robotics Applications}
\label{app:robotics}

\begin{figure}[!t]
  \centering
  \includegraphics[width=0.94\textwidth]{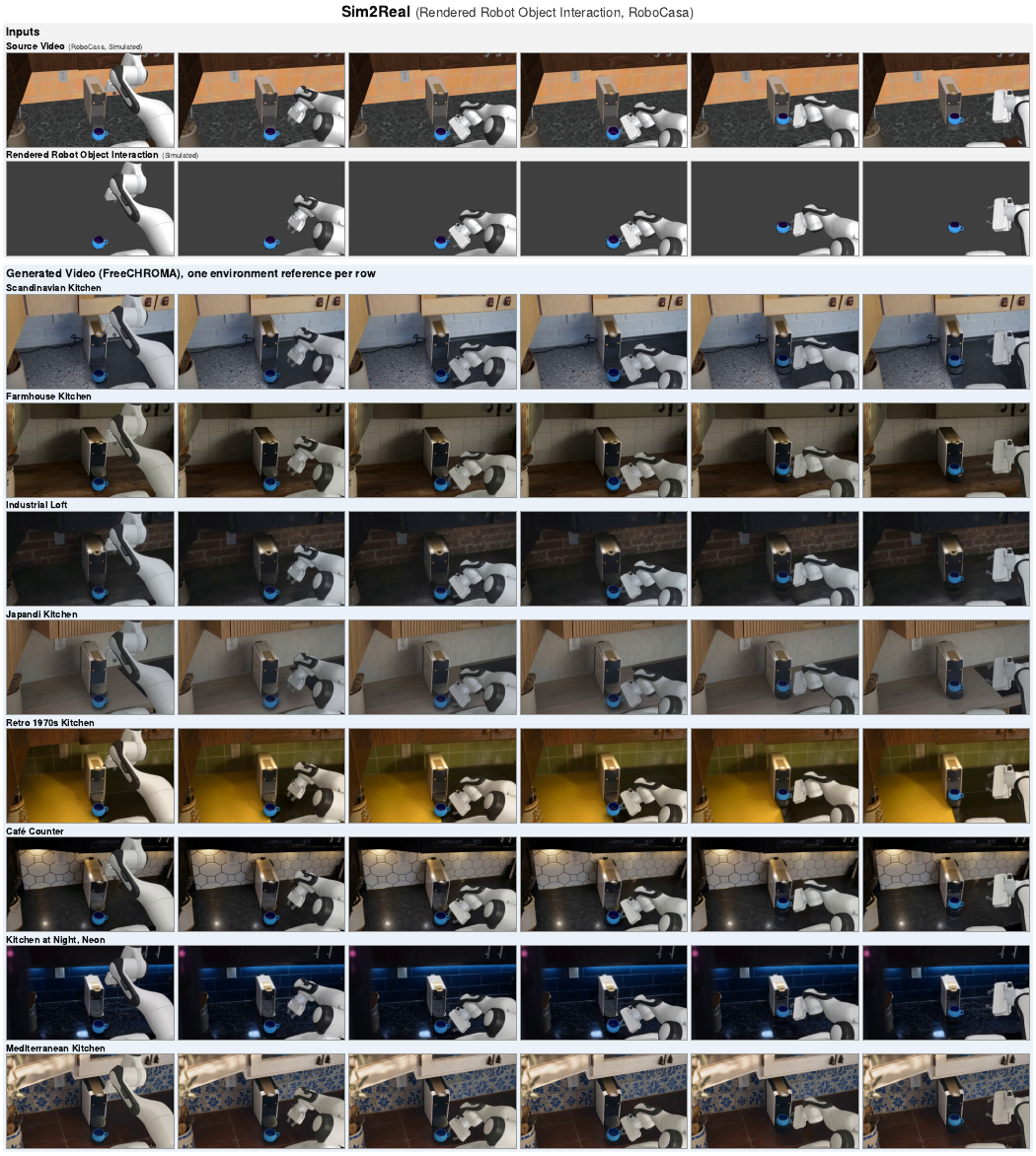}
  \caption{\textbf{Sim2Real on RoboCasa~\citep{robocasa365}: key frames.} Top: the simulated source video and the rendered robot-object interaction used as foreground control (the arm together with the manipulated objects). Below: the videos generated by FreeCHROMA from eight different environment reference images, shown at the same frames. The same simulated action is re-rendered into every environment while the arm motion and the object interaction are preserved. \textcolor{blue}{See the \href{\projecturl}{project page} for more video examples.}}
  \label{fig:sim2real_grid}
\end{figure}

We explore robotics applications using the same generation pipeline without robot-specific training. A masked foreground sequence specifies the scene elements and motions to preserve, while background references and task prompts guide the generated environment and interactions.

\paragraph{Sim-to-real transfer.}
We use rendered sequences from RoboCasa~\citep{robocasa365}, retaining the robot body and interacting objects, together with their motion, as foreground conditions. A single background reference specifies the target environment, and a task prompt provides interaction guidance. \method{} synthesizes the surrounding scene and harmonizes the retained foreground with the reference environment (Fig.~\ref{fig:sim2real_grid}).
\textcolor{blue}{See the \href{\projecturl}{project page} for more video examples.}

\begin{figure}[!t]
  \centering
  \includegraphics[width=\textwidth]{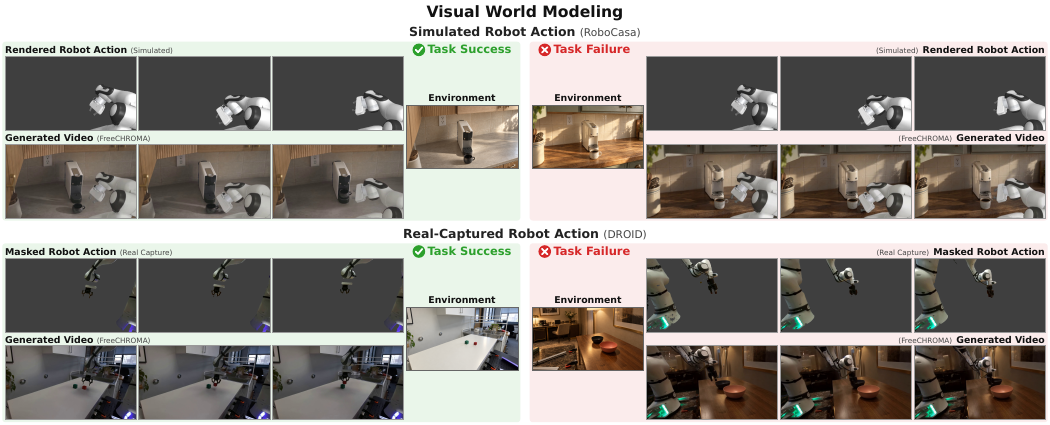}
  \caption{\textbf{Qualitative robotics applications.} Rendered RoboCasa~\citep{robocasa365} (top) and masked DROID~\citep{khazatsky2024droid} (bottom) arms; green/red panels show successful/failed interactions. \textcolor{blue}{See the \href{\projecturl}{project page} for more video examples.}}
  \label{fig:robot}
\end{figure}

\begin{samepage}
\paragraph{Visual world modeling.}
We render the robot body from RoboCasa~\citep{robocasa365} sequences and segment it from DROID~\citep{khazatsky2024droid} sequences, retaining only the robot action and excluding interacting objects.
The input sequence specifies the robot action, while the model generates the surrounding scene and the objects' responses to that action, conditioned on the reference environment and task prompt (Fig.~\ref{fig:robot}).
\textcolor{blue}{See the \href{\projecturl}{project page} for more video examples.}
\par
\end{samepage}

\section{VLM Evaluation Protocol}
\label{app:vlm}

We use Qwen3-VL-8B-Instruct to rate how well each generated video preserves the subject and places it in the new scene. The rating criteria are adapted from PAI-Studio~\citep{gao2026paistudiocinematicvideobackground}. For each video, the model sees 12 frames spaced evenly from beginning to end. It gives separate scores for three aspects of the result: whether the subject’s brightness, light direction, color, and shadows fit the background throughout the video; whether the subject has a believable size, position, and relationship to its surroundings rather than appearing to float; and whether its outlines remain clean, including difficult areas such as hair, fingers, and motion-blurred edges. We assess subject preservation separately. For that rating, the model sees 12 evenly spaced frames from the original foreground-only video and 12 from the generated result, and checks whether the subject’s identity, shape, clothing, pose, and movement have been retained. The instructions explicitly allow lighting and color to change to suit the new scene. The model is not shown the target background reference. Each score is an integer from 1 to 10 and must include a brief explanation. We repeat the ratings three times with repeatable settings (temperature 0.2 and top-p 0.9), take the median of the three scores for each generated video, and report the average over 1,000 generated videos (500 foreground videos, each paired with two distinct background conditions).
\section{Details of Data Construction Pipeline}
\label{app:data}

We construct pseudo-pairs from natural videos through the automated pipeline in \cref{sec:data-pipeline}. Foreground appearance and background references are augmented, while the original video $V$ remains the target.

\paragraph{Video selection and foreground extraction.}
We retain short clips with a clear foreground, sufficient visible background, and no shot transitions. We use \href{https://www.anthropic.com/news/claude-opus-4-6}{Claude Opus~4.6} as the vision-language model (VLM) for foreground-entity identification and visual quality checks. SAM~3.1~\citep{carion2026sam} segments and tracks the subject and interacting objects; their masks are merged into a joint matte $M$ to extract $F=M\odot V$. Clips with unstable masks, severe blur, or excessive foreground occupancy are rejected.

\paragraph{Normal estimation.}
We use \href{https://normalcrafter.github.io/}{NormalCrafter} to estimate video surface normals, aligned with the foreground frames for the illumination augmentation in Appendix~\ref{app:illum-aug}.

\paragraph{Background references.}
\label{app:data_bg}
We use camera trajectories estimated by ViPE to assess motion magnitude and select one to three source keyframes accordingly. \href{https://huggingface.co/tencent/HunyuanImage-3.0-Instruct}{HunyuanImage-3.0-Instruct} removes the selected foreground entities from these keyframes. \href{https://huggingface.co/black-forest-labs/FLUX.2-klein-9B}{FLUX.2-klein-9B} then changes the viewpoint and crop to produce unaligned background references. The VLM rejects candidates that alter scene identity or introduce conspicuous artifacts.

\subsection{Foreground illumination augmentation}
\label{app:illum-aug}

We perturb the foreground condition to prevent direct copying of its original illumination. The target video remains unchanged, so the model must recover a compatible appearance while preserving the performance.

\paragraph{Photometric changes.}
We sample $s\sim\mathcal U(0.3,1.4)$, replacing it with a mild draw from $\mathcal U(0,0.1)$ for $5\%$ of clips. White balance, gamma, and saturation use $s_c=0.35s$; the remaining operators use $s$ (\cref{tab:illum-parameters}). Operators are applied in approximately linear light, $\ell=x^{2.2}$, followed by clipping and conversion back to display encoding. All random parameters are shared across the clip to avoid artificial flicker.

\begin{table}[!htbp]
    \centering
    \caption{Photometric augmentation, applied in the listed order. Each occurrence of $u$ is sampled independently from $\mathcal U(-1,1)$.}
    \label{tab:illum-parameters}
    \small
    \begin{tabular}{@{}p{0.24\linewidth}p{0.69\linewidth}@{}}
        \toprule
        Operator & Sampling rule \\
        \midrule
        Exposure & $\log g\sim\mathcal N(0,(0.40s)^2)$ \\
        White balance & Channel log gains $\sim\mathcal N(0,(0.25s_c)^2)$; gains normalized by their mean \\
        Contrast & $q=1+0.40su$, about linear mid-gray $0.18$ \\
        Gamma & $\log\kappa_j\sim\mathcal N(0,(0.20s_c)^2)$ for each channel $j$ \\
        Saturation & $\sigma=1+0.50s_c u$, about BT.601 luma \\
        \bottomrule
    \end{tabular}
\end{table}

With probability $0.6$, we also apply a fixed spatial field combining a directional ramp and radial vignette. The ramp direction is uniform on $[0,2\pi)$, its amplitude lies in $[0,0.5s]$, and the vignette strength lies in $[0,0.4s]$. The combined gain is clipped to $[0.2,2.0]$.

\paragraph{Normal-guided shading.}
When frame-aligned normals are available, they are resized with the foreground and renormalized. We sample $L\sim\mathcal N(0,I_3)$, replace $L_z$ with $|L_z|+0.3$, and normalize it. The shading and relative gain are
\begin{align}
    S &= a+k\max(0,\widehat N^\top L),
    \label{eq:lambertian-aug}\\
    \rho &= \operatorname{clip}
    \!\left(1+0.6s\left(S/\overline S-1\right),\,0.3,\,2.5\right),
    \label{eq:relight-ratio}
\end{align}
where $a\sim\mathcal U(0.4,0.8)$, $k\sim\mathcal U(0.3,0.9)$, and $\overline S$ is the foreground mean. This normalization separates directional shading from global exposure. The light parameters remain fixed throughout the clip; unavailable normals fall back to photometric augmentation.

\paragraph{Foreground support.}
For inputs in $[0,1]$, the augmentation uses
$\widehat M_f(p)=\mathbf{1}[\max_j F_f(j,p)>8/255]$
to recover foreground support. The final condition is
\begin{equation}
    T_\theta(F;N)=\widehat M\odot
    \Psi_{\mathrm{normal}}\!\left(\Psi_{\mathrm{phot}}(F);N\right),
    \label{eq:illum-aug-full}
\end{equation}
where the two operators denote the photometric and optional shading branches. Multiplication by $\widehat M$ restores the black background. These transforms provide approximate illumination mismatches; the natural video supplies the harmonized target.

\section{Datasets and Implementation Details}
\label{app:implementation}
We collect an initial pool of 453K video clips from HumanVid-Real~\citep{DBLP:conf/nips/00010ZF0LTCX0L24} and OpenVidHD~\citep{nan2024openvid}. Following the filtering pipeline in Fig.~\ref{fig:data-pipeline}, we apply coarse filtering based on foreground occupancy, background clarity, temporal continuity, and the presence of subtitles, logos, or black borders, retaining 45K high-quality clips. We then rebalance the pool by motion level and the numbers of foreground people and objects to obtain a final set of 35K clips, of which 34.1K are used for training and the remaining 900 clips are held out for validation and testing.

We initialize from Wan2.2-Fun-A14B-Control~\citep{wan2025} and fine-tune its high- and low-noise experts separately, using a nominal noise boundary of $0.9$ and scheduler shift $5.0$. Each run uses eight NVIDIA H200 GPUs with BF16 precision and gradient checkpointing. Training clips contain 81 frames at $720\times1280$ resolution. We use one clip per GPU without gradient accumulation, giving a global batch size of eight. The final model uses the checkpoint saved at step 4,500 from each expert.

We apply LoRA~\citep{hu2022lora} with rank $64$ and $\alpha=32$ to the attention query, key, and value projections and the linear layers of feed-forward networks. In the input projection, we freeze the weights corresponding to the noisy video latent and train only those corresponding to the foreground and mask conditions. We also train the background-reference convolution. The VAE, text encoder, and remaining backbone weights are frozen. AdamW uses learning rates of $10^{-4}$ for LoRA and $10^{-5}$ for the input layers and background-reference convolution, with weight decay $0.03$ and $0$, respectively. Both groups use cosine decay with 50 warmup steps and gradient-norm clipping at $0.05$.

Each example provides a masked foreground video, its binary mask sequence, and up to three background references. Text dropout is $10\%$. Background conditioning uses $10\%$ whole-set dropout and $10\%$ per-reference dropout, retaining at least one valid view when the set is kept.

At inference, we use 25 sampling steps with a classifier-free guidance scale of 4.5.